\documentclass[11pt,a4paper]{article}

\usepackage[utf8]{inputenc}
\usepackage[T1]{fontenc}
\usepackage{lmodern}
\usepackage[margin=1in]{geometry}
\usepackage{amsmath,amssymb,amsthm}
\usepackage{mathtools}
\usepackage{booktabs}
\usepackage{graphicx}
\usepackage{xcolor}
\usepackage{hyperref}
\usepackage{cleveref}
\usepackage{caption}
\usepackage{subcaption}
\usepackage{tikz}
\usepackage{pgfplots}
\usepackage{algorithm}
\usepackage{algpseudocode}
\usepackage{float}
\usepackage{enumitem}
\usepackage{microtype}
\usepackage{natbib}
\usepackage{mdframed}

\pgfplotsset{compat=1.17}
\usetikzlibrary{patterns, arrows.meta, positioning, calc, shapes.geometric, fit, shadows}

\hypersetup{
    colorlinks=true,
    linkcolor={blue!70!black},
    citecolor={green!50!black},
    urlcolor={blue!80!black},
    pdftitle={SFT-GRPO Data Overlap as a Post-Training Hyperparameter for Autoformalization},
    pdfauthor={},
}

\definecolor{osmosisblue}{HTML}{2563EB}
\definecolor{osmosisgray}{HTML}{64748B}
\definecolor{osmosisgreen}{HTML}{16A34A}
\definecolor{osmosisred}{HTML}{DC2626}
\definecolor{codebg}{HTML}{F8FAFC}
\definecolor{codeframe}{HTML}{E2E8F0}

\usepackage{listings}
\lstdefinelanguage{Lean4}{
  keywords={theorem, lemma, def, let, in, where, have, show, by, sorry,
            import, open, namespace, end, section, variable, instance,
            class, structure, inductive, match, with, if, then, else,
            return, do, for, fun, example, noncomputable},
  keywordstyle=\color{blue!60!black}\bfseries,
  sensitive=true,
  comment=[l]{--},
  morecomment=[s]{/-}{-/},
  commentstyle=\color{osmosisgray}\itshape,
  stringstyle=\color{osmosisgreen},
  morestring=[b]",
}
\title{SFT-GRPO Data Overlap as a Post-Training Hyperparameter\\for Autoformalization}
\author{
  \textsc{Xiaole Su} \quad \textsc{Kasey Zhang} \quad \textsc{Andy Lyu} \\[4pt]
  Osmosis AI \\[2pt]
  {\small \texttt{xiaole@osmosis.ai} \quad \texttt{kasey@osmosis.ai} \quad \texttt{andy@osmosis.ai}}
}
\date{}

\begin{document}
\maketitle

\begin{abstract}
Supervised fine-tuning (SFT) followed by Group Relative Policy Optimization (GRPO) is a common post-training recipe. We conduct a controlled ablation over SFT-GRPO data overlap, evaluating Qwen3-8B (thinking disabled) post-trained for Lean~4 autoformalization under six conditions that differ solely in training recipe: a base model, SFT-only, GRPO-only, and three SFT+GRPO configurations where 0\%, 30\%, or 100\% of the GRPO prompts coincide with the SFT corpus. Keeping SFT and GRPO data disjoint consistently outperforms full overlap at zero additional compute cost. Evaluating on Gaokao-Formal and PutnamBench under both compile pass@$k$ and semantic pass@$k$ (assessed by an LLM judge), we find that: (1)~lower overlap is monotonically associated with higher compilation and semantic accuracy; (2)~at 0\% overlap, GRPO yields a 10.4 percentage-point (pp) semantic gain over SFT alone on Gaokao; (3)~at 100\% overlap, both compilation and semantic accuracy remain flat, rendering the GRPO stage effectively redundant; (4)~dual-metric evaluation reveals compile-semantic gaps exceeding 30~pp for the highest-compiling models, a disparity invisible under compile-only benchmarking. To our knowledge, this is the first controlled investigation of SFT-GRPO data overlap as a post-training hyperparameter, showing how model behavior varies based on the degree of data sharing between stages.
\end{abstract}

\section{Introduction}\label{sec:intro}

Autoformalization, the translation of informal mathematics into machine-verifiable formal languages, is a prerequisite for scaling formal verification to the breadth of human mathematical knowledge \citep{wu2022autoformalization}. Lean~4 \citep{demoura2021lean4} and its mathematical library Mathlib have emerged as the dominant target, with recent systems achieving high compilation rates on standard benchmarks \citep{wang2025kimina, zhang2025mathesis, wu2025stepfun}.

Prevailing approaches combine supervised fine-tuning (SFT) on labeled (natural-language, Lean~4) pairs with reinforcement learning, typically Group Relative Policy Optimization (GRPO) \citep{shao2024deepseekmath}, using compiler feedback as part of the reward signal. While the SFT-GRPO pipeline has become standard practice, a fundamental design decision remains underexplored: most published pipelines draw their GRPO corpus from the same pool as the SFT data (or use identical data) with no explicit consideration of overlap. We investigate how the degree of training data overlap between the SFT and GRPO stages affects downstream formalization quality. Concurrently, compilation pass rate remains the predominant evaluation metric in autoformalization studies \citep{wang2025kimina, huang2025formalrl}, often leaving open whether compiling outputs faithfully capture the semantics of the original mathematical statement.

\begin{figure}[t]
\centering
\begin{mdframed}[backgroundcolor=codebg, linecolor=codeframe, linewidth=0.6pt, roundcorner=2pt]
{\small
\textbf{Natural-language problem (Putnam 1962 A6):}\\[2pt]
Let $S$ be a set of rational numbers closed under addition and multiplication, with the property that for every rational $r$ exactly one of $r \in S$, $-r \in S$, $r = 0$ holds. Prove that $S$ is the set of all positive rationals.\\[6pt]
\textbf{Lean~4 formalization:}
\begin{lstlisting}
theorem putnam_1962_a6 (S : Set ℚ)
  (hSadd : ∀ a ∈ S, ∀ b ∈ S, a + b ∈ S)
  (hSprod : ∀ a ∈ S, ∀ b ∈ S, a * b ∈ S)
  (hScond : ∀ r : ℚ, (r ∈ S ∨ -r ∈ S ∨ r = 0)
    ∧ ¬(r ∈ S ∧ -r ∈ S)
    ∧ ¬(r ∈ S ∧ r = 0)
    ∧ ¬(-r ∈ S ∧ r = 0))
  : S = { r : ℚ | r > 0 } := sorry
\end{lstlisting}
}
\end{mdframed}
\caption{The autoformalization task: translate a natural-language mathematical statement into a Lean~4 theorem declaration that typechecks against Mathlib. The model must encode the problem's hypotheses and conclusion as typed Lean~4 expressions; \texttt{sorry} stands in for the proof (\S\ref{sec:task}).}
\label{fig:task_example}
\end{figure}

This paper reports a controlled study of SFT-GRPO data overlap as a post-training hyperparameter for Lean~4 autoformalization. Holding the base model (Qwen3-8B, thinking disabled), data sources, and all other hyperparameters fixed, we evaluate six conditions differing solely in training recipe and in the fraction of GRPO prompts that overlap with the SFT corpus (0\%, 30\%, 100\%). We evaluate on Gaokao-Formal \citep{zhang2024gaokao} (495 problems, moderate difficulty) and PutnamBench \citep{tsoukalas2024putnambench} (672 problems, high difficulty) using both compile pass@$k$ and semantic pass@$k$, where the latter quantifies semantic faithfulness using an LLM judge (threshold $\geq 0.7$).

Our principal contributions are:
\begin{enumerate}[leftmargin=*, itemsep=3pt]
  \item \textbf{Controlled ablation of SFT-GRPO data overlap.} Lower overlap is monotonically associated with higher compilation and semantic accuracy. At 0\% overlap, GRPO yields a 10.4~pp semantic gain over SFT alone on Gaokao. At 100\% overlap, both metrics remain flat, indicating that non-overlapping data is a necessary condition for GRPO to confer meaningful benefit.

  \item \textbf{Dual-metric evaluation.} We evaluate using both compile pass@$k$ and semantic pass@$k$, revealing compile-semantic gaps exceeding 30~pp for the highest-compiling models. These gaps are invisible under compile-only evaluation.
\end{enumerate}

\section{Background and Related Work}\label{sec:related}

\paragraph{Autoformalization.}
Autoformalization encompasses the full translation of informal mathematics into formal, machine-verifiable languages, covering both problem statements and proofs \citep{wu2022autoformalization, weng2025autoformalization}. Within this broad endeavor, \emph{statement formalization} is a distinct subtask: translating a natural-language theorem statement into a formal declaration without producing a proof. Accurate statement formalization serves as a critical starting point for downstream \emph{prover models} such as Kimina-Prover \citep{wang2025kimina}, DeepSeek-Prover-V2 \citep{xin2025deepseekprover}, and Goedel-Prover \citep{lin2025goedelprover}, which handle tactic generation and proof search.

Recent training approaches for statement formalization span a broad spectrum: SFT on distilled data \citep{wang2025kimina}, SFT combined with RL under various reward designs \citep{zhang2025mathesis, wu2025stepfun}, pure RL without labeled data \citep{huang2025formalrl}, and retrieval-augmented or chain-of-thought prompting strategies \citep{weng2025autoformalization}. A key design axis across these systems is the reward signal for RL. FormaRL \citep{huang2025formalrl} uses a binary reward combining compilation success and LLM consistency checking. StepFun \citep{wu2025stepfun} employs Bidirectional Extended Definitional Equivalence (BEq) as a formally grounded reward. Mathesis \citep{zhang2025mathesis} introduces LeanScorer for partial-credit feedback. Our GRPO configuration is closest to FormaRL's but augments the reward with a continuous semantic judge score rather than relying solely on binary feedback.

\paragraph{SFT-RL interaction.}
DeepSeek-R1 \citep{guo2025deepseekr1} demonstrates that RL can generalize beyond the SFT training distribution, and \citet{chu2025sftmemorizes} provide complementary evidence, showing that SFT tends to memorize training distributions while RL generalizes beyond them. \citet{lu2026datacentric} advance a data-centric perspective on RLHF, arguing that RL's advantage derives partly from implicit data curation through the reward signal. Our work extends this line of inquiry by isolating a specific data-management variable, the overlap between SFT and GRPO pools, and quantifying its effect on both compilation and semantic accuracy.

\paragraph{Evaluation beyond compilation.}
Most autoformalization work evaluates only whether generated statements typecheck \citep{wang2025kimina, huang2025formalrl}. BEq provides the strongest semantic guarantee available, formally verifying logical equivalence through bidirectional proof construction, though it suffers from high false-negative rates when the prover cannot close equivalent but syntactically dissimilar goals \citep{wu2025stepfun, weng2025autoformalization}. LeanScorer offers a more granular partial-credit assessment \citep{zhang2025mathesis}. LLM-judge approaches trade formal guarantees for scalability and coverage; \citet{weng2025autoformalization} identify semantic evaluation as an open challenge for the field. We adopt an LLM judge as a scalable proxy (\S\ref{sec:eval}).

\section{Data}\label{sec:data}

\subsection{Task Definition}\label{sec:task}

We study \emph{autoformalization}: translating natural-language mathematical statements into Lean~4 theorem declarations that typecheck against Mathlib~v4.27.0. Each output is a standalone theorem terminated by \texttt{:= by sorry}, where \texttt{sorry} is a Lean~4 keyword that accepts any goal without a proof, allowing the statement to typecheck independently of proof search. No proof body is required; this formulation isolates statement formalization from proving.

\subsection{Sources and Processing}\label{sec:sources}

We draw from four publicly available corpora: \textbf{NuminaMath} \citep{li2024numinamath} (104K competition math problems), \textbf{Leanabell-Prover} \citep{stoney2025leanabell} (1.13M aggregated entries), \textbf{HERALD} \citep{gao2025herald} (580K NL--FL pairs from Mathlib4), and \textbf{Lean Workbook} \citep{he2024leanworkbook} (13.5K contest-style pairs). Each source was vetted by compiling a 500-sample slice against Mathlib~v4.27.0 (\Cref{tab:source_quality}).

\begin{table}[t]
\centering
\caption{Source quality: 500-sample compilation test against Mathlib~v4.27.0.}
\label{tab:source_quality}
\begin{tabular}{@{}lrc@{}}
\toprule
\textbf{Dataset} & \textbf{Pool Size} & \textbf{Pass Rate} \\
\midrule
Leanabell-Prover \citep{stoney2025leanabell} & 1.13M & 93.0\% \\
Lean Workbook \citep{he2024leanworkbook} & 13.5K & 92.0\% \\
NuminaMath \citep{li2024numinamath} & 104K & 90.6\% \\
HERALD \citep{gao2025herald} & 580K & 86.6\% \\
\bottomrule
\end{tabular}
\end{table}

The curation pipeline comprises seven stages: ingestion, compilation (against Mathlib~v4.27.0), deduplication (SHA-256, essential because Leanabell aggregates 15--20\% duplicate content from NuminaMath and Lean Workbook), quality filtering, topic stratification, answer injection (\S\ref{sec:answer_injection}), and formatting to Alpaca JSONL.

\subsection{SFT Dataset}\label{sec:sft_data}

The SFT dataset consists of 20{,}000 (natural-language, Lean~4) pairs, composed as shown in \Cref{tab:sft_mix}.

\begin{table}[t]
\centering
\caption{SFT dataset (20K) composition.}
\label{tab:sft_mix}
\begin{tabular}{@{}lrcl@{}}
\toprule
\textbf{Source} & \textbf{Count} & \textbf{\%} & \textbf{Role} \\
\midrule
NuminaMath & 8{,}000 & 40\% & Topic-balanced competition math \\
Leanabell & 7{,}000 & 35\% & Olympiad + AoPS diversity \\
HERALD & 3{,}000 & 15\% & Broad Mathlib coverage \\
Lean Workbook & 2{,}000 & 10\% & Proved + disproved statements \\
\midrule
Total & 20{,}000 & 100\% & \\
\bottomrule
\end{tabular}
\end{table}

\subsection{GRPO Dataset and Overlap Definition}\label{sec:grpo_data}

The GRPO corpus consists of ${\sim}$16{,}000 natural-language prompts sampled from the same four sources. Each prompt is paired with a ground-truth Lean~4 statement consumed exclusively by the reward function; the model never observes it during generation. We define \emph{overlap} as the fraction of GRPO prompts that also appeared in the SFT training set:

\begin{itemize}[leftmargin=*, itemsep=2pt]
  \item \textbf{0\% overlap}: all 16K prompts are fresh samples, cross-deduplicated against the SFT pool.
  \item \textbf{30\% overlap}: ${\sim}$4.8K prompts are resampled from the SFT pool; the remaining ${\sim}$11.2K are fresh.
  \item \textbf{100\% overlap}: the entire GRPO corpus is drawn from the SFT pool.
\end{itemize}

All three pools share the same source distribution and differ only in their intersection with the SFT set, isolating overlap as the sole experimental variable. Because the pools are drawn from the same four sources with comparable quality profiles (\Cref{tab:source_quality}), we expect difficulty distributions to be broadly similar across overlap conditions, though we do not formally verify this.

\subsection{Answer Injection}\label{sec:answer_injection}

Many competition problems require the model to ``find'' or ``determine'' a value that appears only in the ground-truth Lean~4 statement. A formalization model is not expected to solve the underlying math problem; its task is translation, not computation. However, during GRPO the model must produce a complete formalization from the prompt alone. Without the target value, the model cannot formalize a correct statement because it simply does not know the correct answer. We observe that GRPO-only training without answer injection produces consistently poor formalizations for such problems.

\Cref{fig:answer_injection} illustrates this with a concrete example from LeanWorkbook. Without injection, the model must independently compute that the series $\sum_{k=1}^{\infty} k^2/2^k$ converges to~6 before it can even begin formalizing. Our answer-injection pipeline extracts concrete answers from the Lean~4 ground truth using 12 pattern types and appends them to ``find''-type prompts (e.g., ``Show that the answer is~6''), so the model can focus entirely on formalization rather than re-deriving the answer. This pipeline is particularly relevant for ``find the value'' and ``determine the answer'' problems, which form a substantial fraction of competition mathematics datasets. Most existing autoformalization datasets are designed for SFT with ground-truth outputs, not for RL where the model generates from prompts alone. For problems of this type, answer injection bridges the gap, enabling GRPO training on standard SFT datasets without requiring the model to solve the underlying mathematics.

\begin{figure}[t]
\centering
\begin{mdframed}[backgroundcolor=codebg, linecolor=codeframe, linewidth=0.6pt, roundcorner=2pt]
{\small
\textbf{Without answer injection:}\\[2pt]
\texttt{Prompt:} ``Determine the value of the infinite series $\sum_{k=1}^{\infty} \frac{k^2}{2^k}$.''\\
\texttt{Model must:} (1) compute $\sum k^2/2^k = 6$, then (2) formalize as Lean~4.\\[4pt]
\textbf{With answer injection:}\\[2pt]
\texttt{Prompt:} ``Determine the value of the infinite series $\sum_{k=1}^{\infty} \frac{k^2}{2^k}$. Show that the answer is 6.''\\
\texttt{Model must:} (1) formalize the statement directly.\\[4pt]
\textbf{Lean 4 output:}
\begin{lstlisting}
theorem lean_workbook :
  (*$\sum'$*) k : (*$\mathbb{N}$*), (k ^ 2 : (*$\mathbb{R}$*)) / 2 ^ k = 6 := by sorry
\end{lstlisting}
}
\end{mdframed}
\caption{Answer injection eliminates the need for the model to solve the underlying math problem during GRPO. Without the injected answer, the model does not know the correct value and cannot formalize a correct statement.}
\label{fig:answer_injection}
\end{figure}

\section{Training}\label{sec:training}

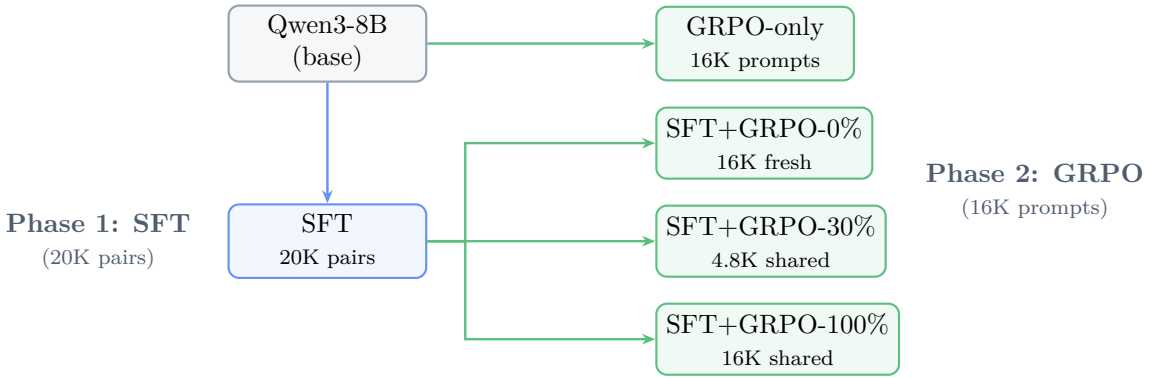
\begin{figure}[t]
\centering
\begin{tikzpicture}[
    node distance=1.6cm and 2.0cm,
    box/.style={rectangle, draw, rounded corners=4pt, minimum width=2.6cm, minimum height=0.85cm,
                align=center, font=\small, line width=0.8pt,
                drop shadow={shadow xshift=0.4pt, shadow yshift=-0.4pt, fill=black!12}},
    basebox/.style={box, draw=osmosisgray!70, fill=osmosisgray!6},
    sftbox/.style={box, draw=osmosisblue!70, fill=osmosisblue!6},
    grpobox/.style={box, draw=osmosisgreen!70, fill=osmosisgreen!6},
    arrow/.style={-{Stealth[length=5pt, width=4pt]}, line width=0.9pt},
    phaselabel/.style={font=\small\bfseries, text=osmosisgray!80!black, align=center},
]
  \node[basebox] (base) {Qwen3-8B\\(base)};

  \node[sftbox, below=of base] (sft) {SFT\\{\scriptsize 20K pairs}};

  \node[grpobox, right=3.0cm of base] (grponly) {GRPO-only\\{\scriptsize 16K prompts}};

  \node[grpobox, right=3.0cm of sft, yshift=1.3cm] (grpo0) {SFT+GRPO-0\%\\{\scriptsize 16K fresh}};
  \node[grpobox, right=3.0cm of sft, yshift=0.0cm] (grpo30) {SFT+GRPO-30\%\\{\scriptsize 4.8K shared}};
  \node[grpobox, right=3.0cm of sft, yshift=-1.3cm] (grpo100) {SFT+GRPO-100\%\\{\scriptsize 16K shared}};

  \draw[arrow, osmosisblue!70] (base) -- (sft);
  \draw[arrow, osmosisgreen!70] (base.east) -- (grponly.west);
  \draw[arrow, osmosisgreen!70] (sft.east) -- ++(0.5,0) |- (grpo0.west);
  \draw[arrow, osmosisgreen!70] (sft.east) -- ++(0.5,0) |- (grpo30.west);
  \draw[arrow, osmosisgreen!70] (sft.east) -- ++(0.5,0) |- (grpo100.west);

  \node[phaselabel, left=0.25cm of sft, xshift=-0.1cm] {Phase 1: SFT\\{\scriptsize\normalfont (20K pairs)}};

  \coordinate (midright) at ($(grponly.east)!0.5!(grpo100.east)$);
  \node[phaselabel, right=0.3cm of midright, xshift=0.2cm] {Phase 2: GRPO\\{\scriptsize\normalfont (16K prompts)}};
\end{tikzpicture}
\caption{Training overview. Phase~1 produces one SFT checkpoint. Phase~2 applies GRPO to both the base model (producing GRPO-only) and the SFT checkpoint (producing three overlap variants). All Phase~2 runs share identical hyperparameters; only the data pool and initialization differ.}
\label{fig:training_overview}
\end{figure}

We use Qwen3-8B \citep{qwen2025qwen3} as the base model with thinking mode \textbf{disabled} (i.e., \texttt{enable\_thinking=False} in the Qwen3 chat template, suppressing internal chain-of-thought generation) for all conditions. No publicly available dataset provides ground-truth reasoning traces for Lean~4 autoformalization, and enabling chain-of-thought (whether natively or through distillation) would introduce an additional confounding variable, preventing attribution of observed gains to the training recipe alone. Training proceeds in two phases, yielding six conditions (\Cref{fig:training_overview}).

\subsection{Phase 1: Supervised Fine-Tuning}\label{sec:sft_training}

Qwen3-8B is fine-tuned on the 20K SFT corpus using Axolotl with full-parameter updates, a cosine learning rate schedule (peak $2 \times 10^{-5}$), and 2 epochs. Full hyperparameters are provided in \Cref{tab:sft_config}.

\subsection{Phase 2: Group Relative Policy Optimization}\label{sec:grpo_training}

GRPO \citep{shao2024deepseekmath} is implemented using Slime/Megatron-LM with 8 rollouts per prompt and a constant learning rate of $1 \times 10^{-6}$. All four GRPO runs (three overlap variants and GRPO-only) share identical hyperparameters; only the initialization (base vs.\ SFT checkpoint) and data pool differ. Full hyperparameters are provided in \Cref{tab:grpo_config}.

\subsection{Reward Function}\label{sec:reward}

\begin{figure}[t]
\centering
\begin{tikzpicture}[
    node distance=1.0cm and 1.8cm,
    box/.style={
      rectangle, draw, rounded corners=4pt, minimum width=3cm, minimum height=0.95cm,
      align=center, font=\small, line width=0.8pt,
      drop shadow={shadow xshift=0.5pt, shadow yshift=-0.5pt, fill=black!12}
    },
    decision/.style={
      diamond, draw, aspect=2.4, inner sep=2pt, align=center, font=\small,
      line width=0.8pt,
      drop shadow={shadow xshift=0.5pt, shadow yshift=-0.5pt, fill=black!12}
    },
    arrow/.style={-{Stealth[length=5pt, width=4pt]}, line width=0.9pt},
    yeslabel/.style={font=\scriptsize\bfseries, text=osmosisgreen!80!black},
    nolabel/.style={font=\scriptsize\bfseries, text=osmosisred!80!black},
]
  \node[box, fill=osmosisblue!6, draw=osmosisblue!70] (input) {Model Output\\(Lean~4 code)};
  \node[decision, below=of input, fill=orange!8, draw=orange!70] (compile) {Compiles?\\{\scriptsize\texttt{lake env lean}}};
  \node[box, below right=0.8cm and 1.5cm of compile, fill=osmosisgreen!6, draw=osmosisgreen!70] (judge) {LLM Judge\\{\scriptsize Gemini Flash 3}};
  \node[box, below=of judge, fill=osmosisgreen!10, draw=osmosisgreen!70] (reward) {$r \in [0, 1]$};
  \node[box, below left=0.8cm and 1.5cm of compile, fill=osmosisred!8, draw=osmosisred!70] (fail) {$r = 0.0$};

  \draw[arrow] (input) -- (compile);
  \draw[arrow] (compile) -- node[yeslabel, right, xshift=2pt, yshift=4pt] {Yes} (judge);
  \draw[arrow] (compile) -- node[nolabel, left, xshift=-2pt, yshift=4pt] {No} (fail);
  \draw[arrow] (judge) -- (reward);
\end{tikzpicture}
\caption{Dual-stage reward function. Compilable outputs are scored by Gemini Flash~3 for semantic faithfulness; non-compiling outputs receive zero reward.}
\label{fig:reward_design}
\end{figure}
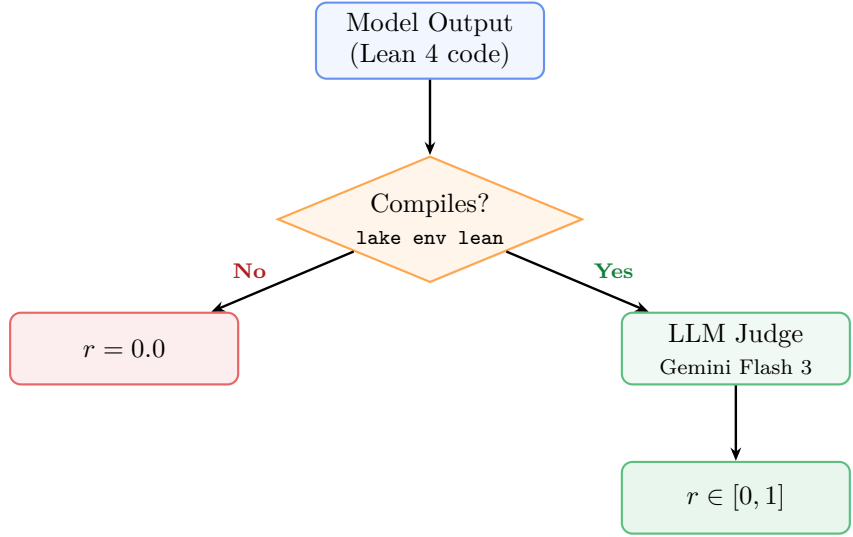

The reward function couples a hard compiler gate with a continuous semantic judge (\Cref{fig:reward_design}, \Cref{alg:reward}). Outputs that fail to typecheck against Mathlib~v4.27.0 receive $r{=}0$. Compiling outputs are scored by Gemini Flash~3, which assesses semantic faithfulness against the ground-truth Lean~4 statement on a continuous scale $r \in [0, 1]$, where $1.0$ denotes a semantically equivalent formalization. This dual-stage design furnishes GRPO with indirect semantic supervision: the model never observes the ground-truth output, yet receives graded feedback on the quality of its own generations.

\begin{algorithm}[t]
\caption{Dual-Stage Reward}\label{alg:reward}
\begin{algorithmic}[1]
\Require NL input $x$, model output $\hat{y}$, ground truth $y^*$
\Ensure Reward $r \in [0,1]$
\State $(ok, \mathit{errors}) \gets \Call{LeanTypeCheck}{\hat{y}, \text{Mathlib v4.27.0}}$
\If{$\neg\, ok$}
    \State \Return $r \gets 0.0$
\EndIf
\State $s_{\text{sem}} \gets \Call{GeminiFlash3Judge}{x,\; \hat{y},\; y^*}$
\State \Return $r \gets s_{\text{sem}}$
\end{algorithmic}
\end{algorithm}

A single system prompt is shared across SFT training, GRPO training, and evaluation (\Cref{app:system_prompt}).

\section{Evaluation}\label{sec:eval}

\subsection{Benchmarks}\label{sec:benchmarks}

We evaluate on two benchmarks spanning distinct difficulty regimes:
\begin{itemize}[leftmargin=*, itemsep=2pt]
  \item \textbf{Gaokao-Formal} \citep{zhang2024gaokao}: 495 problems drawn from the Chinese national college entrance examination (Gaokao), spanning algebra, analysis, and number theory. Moderate difficulty.
  \item \textbf{PutnamBench} \citep{tsoukalas2024putnambench}: 672 problems from the William Lowell Putnam Mathematical Competition, covering a broad range of undergraduate and competition-level mathematics. High difficulty.
\end{itemize}

\subsection{Metrics}\label{sec:metrics_def}

We report two complementary pass@$k$ metrics with $n{=}8$ rollouts per problem:

\begin{itemize}[leftmargin=*, itemsep=2pt]
  \item \textbf{Compile pass@$k$ (C@$k$)}: a rollout counts as a success if it typechecks against Mathlib~v4.27.0.
  \item \textbf{Semantic pass@$k$ (S@$k$)}: a rollout counts as a success only if it compiles \emph{and} receives a semantic faithfulness score $\geq 0.7$ from the judge. This is the primary quality metric.
\end{itemize}

The \textbf{compile-semantic gap} $(\text{C@}k - \text{S@}k)$ measures how often a model produces syntactically valid but semantically incorrect formalizations.

\subsection{Semantic Judge}\label{sec:judge}

Each compiling output is scored by Gemini Flash~3 (temperature 0.0, max 1024 output tokens) on a continuous 0.0--1.0 scale for semantic faithfulness to the ground-truth formalization. Non-compiling outputs receive 0.0. The full judge prompt is provided in \Cref{app:judge_prompt}.

\begin{figure}[t]
\centering
\begin{mdframed}[backgroundcolor=codebg, linecolor=codeframe, linewidth=0.6pt, roundcorner=2pt]
{\small
\textbf{NL problem (Putnam 1962 A6):} Let $S$ be a set of rationals closed under addition and multiplication, with the property that for every rational $r$ exactly one of $r \in S$, $-r \in S$, $r=0$ holds. Prove $S$ is the set of all positive rationals.\\[4pt]
\textbf{Ground truth:}
\begin{lstlisting}
theorem putnam_1962_a6 (S : Set ℚ)
  (hSadd : ∀ a ∈ S, ∀ b ∈ S, a + b ∈ S)
  (hSprod : ∀ a ∈ S, ∀ b ∈ S, a * b ∈ S)
  (hScond : ∀ r : ℚ, (r ∈ S ∨ -r ∈ S ∨ r = 0)
    ∧ ¬(r ∈ S ∧ -r ∈ S)
    ∧ ¬(r ∈ S ∧ r = 0)
    ∧ ¬(-r ∈ S ∧ r = 0))
  : S = { r : ℚ | r > 0 } := sorry
\end{lstlisting}
\textbf{Model output} (compiles, SFT+GRPO-30\%):
\begin{lstlisting}
theorem putnam_1962_a6 (S : Set ℚ)
  (h₁ : ∀ a ∈ S, ∀ b ∈ S, a + b ∈ S ∧ a * b ∈ S)
  (h₂ : ∀ r : ℚ, r ∈ S ∨ -r ∈ S ∨ r = 0)
  : S = {x : ℚ | x > 0} := by sorry
\end{lstlisting}
\textbf{Judge score: 0.5.} ``The candidate captures the additive and multiplicative closure of $S$, and the trichotomy property. However, it incorrectly combines the additive and multiplicative closure into a single hypothesis, and it omits the mutual exclusivity condition in the trichotomy property.''
}
\end{mdframed}
\caption{Semantic judge example. The model output compiles and appears plausible, but omits the mutual-exclusivity constraints ($\neg(r \in S \wedge {-}r \in S)$, etc.), weakening the hypotheses. The judge assigns a partial score of 0.5, reflecting a partially correct formalization.}
\label{fig:judge_example}
\end{figure}

\subsection{External Baselines}\label{sec:baselines_def}

We include two contemporary specialized autoformalizers evaluated under the same harness for contextual comparison: \textbf{Kimina-Autoformalizer-7B} \citep{wang2025kimina} and \textbf{Mathesis-7B} \citep{zhang2025mathesis}.

\subsection{Inference}\label{sec:inference}

All models are served with vLLM on H200 GPUs with $n{=}8$ rollouts per problem. Internal models use temperature 1.0; external baselines use temperature 0.6, following the inference settings recommended on their respective model cards.

\section{Results}\label{sec:results}

\Cref{tab:main_results} reports results across all conditions and benchmarks. SFT and GRPO each individually improve upon the base model; their combination surpasses either in isolation. The remainder of this section focuses on the overlap effect, the central variable of interest.

\begin{table}[t]
\centering
\caption{Compile pass@$k$ (C@$k$) and semantic pass@$k$ (S@$k$) across all model conditions on Gaokao-Formal (495 problems) and PutnamBench (672 problems). $\Delta$: compile-semantic gap (C@1$-$S@1).}
\label{tab:main_results}
\small
\begin{tabular}{@{}l ccccc c ccccc@{}}
\toprule
& \multicolumn{5}{c}{\textbf{Gaokao-Formal}} && \multicolumn{5}{c}{\textbf{PutnamBench}} \\
\cmidrule(lr){2-6} \cmidrule(lr){8-12}
\textbf{Model} & C@1 & C@8 & S@1 & S@8 & $\Delta$ && C@1 & C@8 & S@1 & S@8 & $\Delta$ \\
\midrule
\multicolumn{12}{@{}l}{\textit{Internal models}} \\
Base & 19.9 & 40.2 & 10.2 & 19.4 & 9.7 && 11.3 & 29.3 & 3.3 & 7.7 & 8.0 \\
SFT & 61.8 & 92.3 & 41.0 & 70.9 & 20.8 && 28.5 & 65.6 & 14.3 & 34.2 & 14.2 \\
GRPO-only & 50.9 & 69.3 & 28.1 & 40.2 & 22.8 && 36.1 & 60.1 & 11.9 & 19.2 & 24.2 \\
\addlinespace[3pt]
SFT+GRPO-0\% & 77.6 & 93.9 & 51.4 & 72.7 & 26.2 && 47.9 & 78.1 & 23.6 & 43.0 & 24.3 \\
SFT+GRPO-30\% & 76.4 & 92.9 & 48.6 & 70.7 & 27.8 && 46.4 & 71.9 & 22.9 & 38.8 & 23.5 \\
SFT+GRPO-100\% & 62.9 & 92.7 & 40.6 & 69.9 & 22.3 && 29.1 & 65.3 & 14.7 & 34.8 & 14.4 \\
\addlinespace[3pt]
\multicolumn{12}{@{}l}{\textit{External baselines}} \\
Kimina-7B & 84.2 & 97.2 & 44.6 & 68.1 & 39.6 && 53.5 & 85.7 & 36.8 & 65.3 & 16.7 \\
Mathesis-7B & 84.1 & 95.2 & 49.8 & 71.1 & 34.3 && 63.0 & 88.8 & 43.0 & 69.3 & 20.0 \\
\bottomrule
\end{tabular}
\end{table}

\subsection{The Overlap Effect}\label{sec:overlap_effect}

\begin{figure}[t]
\centering
\includegraphics[width=\textwidth]{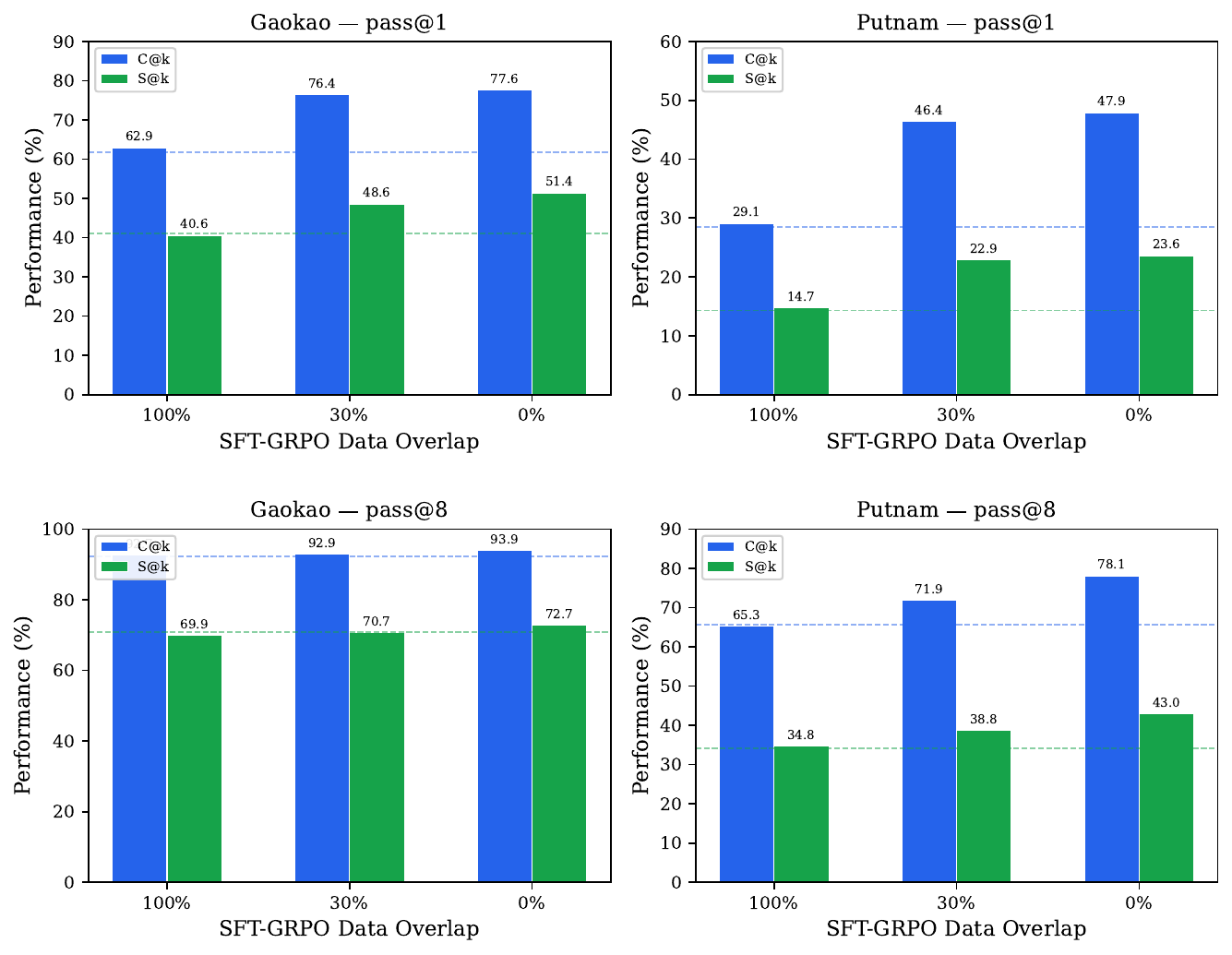}
\caption{Compile and semantic pass@$k$ as a function of SFT-GRPO data overlap. Top: pass@1. Bottom: pass@8. Dashed lines indicate SFT baselines.}
\label{fig:overlap_gradient}
\end{figure}

The overlap gradient is monotonic across both metrics and both benchmarks (\Cref{fig:overlap_gradient}). As a reference point, GRPO-only (RL applied directly to the base model without SFT initialization) improves compilation over the base but trails SFT on semantic accuracy, with a disproportionately large compile-semantic gap (22.8~pp on Gaokao, 24.2~pp on Putnam). This likely reflects the compound effect of limited training duration (the reward curves in \Cref{fig:training_dynamics} suggest training has not fully converged), the compilation hard-gate prioritizing syntax over semantics, and the absence of ground-truth examples that SFT provides.

\paragraph{Non-overlapping data is essential for GRPO efficacy.}
At 0\% overlap, GRPO produces the largest gains: +10.4~pp over SFT alone on Gaokao S@1 and +9.3~pp on Putnam. The model never observes ground-truth outputs during GRPO; the dual-stage reward guides policy exploration toward higher semantic fidelity solely by reward feedback on unseen problems.

\begin{figure}[t]
\centering
\includegraphics[width=\textwidth]{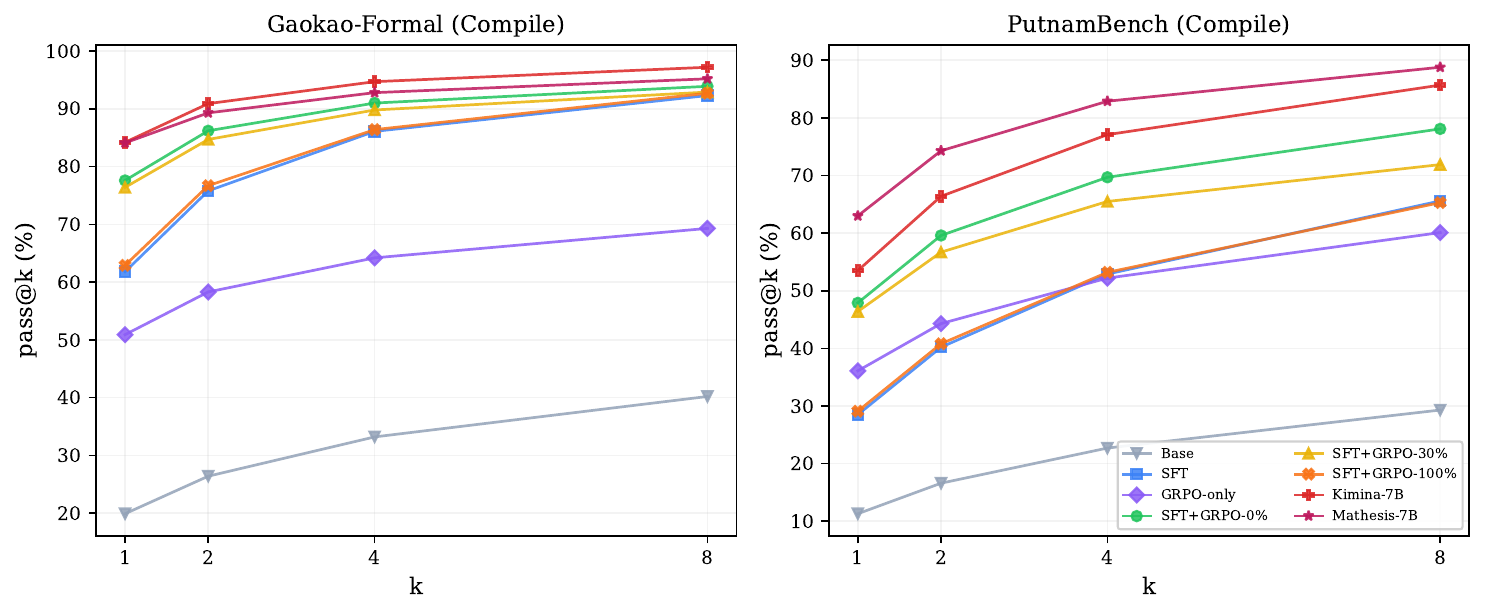}
\caption{Compile pass@$k$ as a function of the number of attempts ($k$) on Gaokao-Formal (left) and PutnamBench (right). Semantic pass@$k$ values at $k{=}1$ and $k{=}8$ are reported in \Cref{tab:main_results} and \Cref{tab:full_semantic}; the overlap ordering is preserved on both metrics.}
\label{fig:passk_curves}
\end{figure}

\paragraph{Full overlap confers no benefit on either metric.}
At 100\% overlap, both C@1 and S@1 remain flat relative to SFT on both benchmarks. The model has likely already acquired formalizations for these problems during SFT; re-encountering them under GRPO yields no additional learning signal. The effect is amplified on harder tasks: the C@1 spread across overlap levels is 18.8~pp on Putnam versus 14.7~pp on Gaokao, and at pass@8, Gaokao variants converge ($\sim$93\%) while Putnam retains a 12.8~pp separation (\Cref{fig:overlap_gradient,fig:passk_curves}).

\subsection{Training Dynamics}\label{sec:dynamics}

\begin{figure}[t]
\centering
\includegraphics[width=\textwidth]{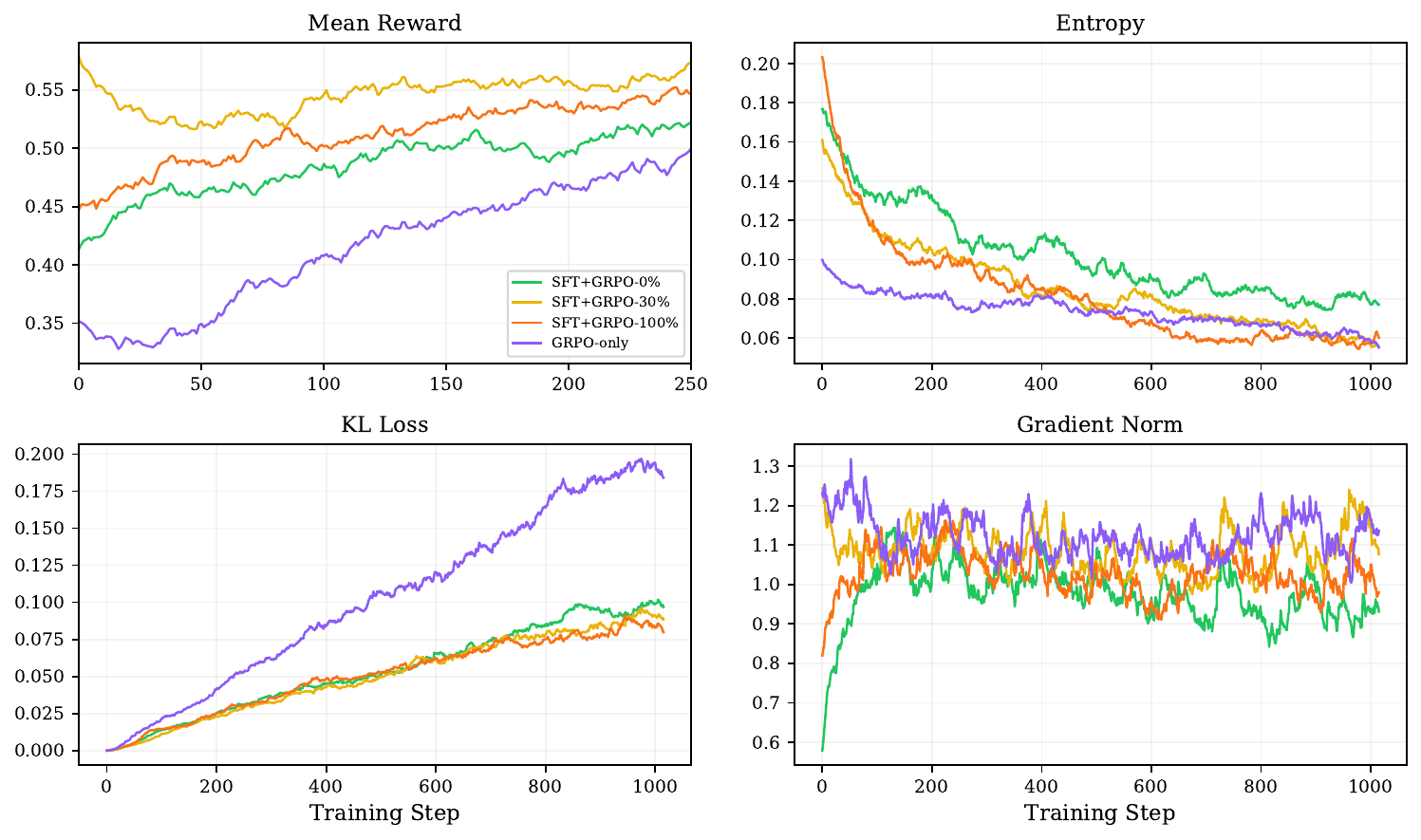}
\caption{Smoothed training dynamics (EMA) across GRPO variants. Higher training reward on SFT+GRPO-30\% does not correspond to better evaluation performance. See \Cref{app:training_dynamics} for raw traces.}
\label{fig:training_dynamics}
\end{figure}

\Cref{fig:training_dynamics} plots training metrics across GRPO runs. SFT+GRPO-30\% attains the highest mean reward yet underperforms SFT+GRPO-0\% at evaluation, indicating that elevated training reward on partially familiar data does not translate into superior generalization.

\subsection{External Baselines and the Compile-Semantic Gap}\label{sec:baselines}

Both external baselines lead on compilation across both benchmarks (Gaokao C@1 ${\sim}$84\% vs.\ our best 77.6\%; Putnam C@1 up to 63.0\%). The compile-semantic gap, however, reveals systematic disparities obscured by compilation alone: models that compile at higher rates, particularly on harder problems, tend to exhibit larger gaps, with Kimina's gap on Gaokao (39.6~pp) and Mathesis's (34.3~pp) being the largest observed. This pattern may partly reflect that successfully compiling harder problems makes preserving full semantic accuracy more challenging, as the formalization must navigate more complex type-theoretic constraints. We also note that the gap is partly a function of the higher compilation ceiling itself: a model with C@1 of 84\% has substantially more room for a compile-semantic gap than one at 48\%, independent of any difficulty effect. Regardless of its cause, this observation motivates dual-metric evaluation: compile-only benchmarking would entirely obscure these disparities. On Putnam, the baselines lead on both metrics, consistent with their larger and more specialized training corpora. Baselines also use temperature 0.6 (per their respective model cards) versus our 1.0, which may influence absolute pass@1 comparisons. Because of this temperature mismatch and differences in training data, comparisons with external baselines are contextual rather than strictly controlled; the primary claims of this paper rest on the within-study overlap ablation, where all internal models share identical decoding settings.

\subsection{Best-of-8 Score Distribution}\label{sec:bestof8}

Best-of-8 scores report the maximum semantic score across 8 rollouts per problem, measuring the oracle upper bound on quality within a fixed rollout budget. This isolates whether high-quality outputs exist in the model's distribution even when single samples are noisy, complementing the pass@$k$ view from \Cref{tab:main_results}. On Gaokao, SFT+GRPO-0\% achieves the highest mean score (0.742) and the highest quality conditioned on solvability (mean$_{\neq 0}$: 0.809). On Putnam, the external baselines lead on both metrics.

\begin{table}[t]
\centering
\caption{Best-of-8 semantic score summary. Mean$_{\neq 0}$: average conditioned on at least one rollout compiling.}
\label{tab:bestof8}
\small
\begin{tabular}{@{}l cc c cc@{}}
\toprule
& \multicolumn{2}{c}{\textbf{Gaokao-Formal}} && \multicolumn{2}{c}{\textbf{PutnamBench}} \\
\cmidrule(lr){2-3} \cmidrule(lr){5-6}
\textbf{Model} & Mean & Mean$_{\neq 0}$ && Mean & Mean$_{\neq 0}$ \\
\midrule
Base & 0.215 & 0.678 && 0.106 & 0.563 \\
SFT & 0.719 & 0.805 && 0.375 & 0.697 \\
GRPO-only & 0.423 & 0.712 && 0.244 & 0.576 \\
SFT+GRPO-0\% & 0.742 & 0.809 && 0.468 & 0.694 \\
SFT+GRPO-30\% & 0.716 & 0.791 && 0.427 & 0.693 \\
SFT+GRPO-100\% & 0.716 & 0.804 && 0.380 & 0.697 \\
\addlinespace[5pt]
\multicolumn{6}{@{}l}{\textit{External baselines}} \\
Kimina-7B & 0.731 & 0.768 && 0.677 & 0.823 \\
Mathesis-7B & 0.732 & 0.785 && 0.719 & 0.820 \\
\bottomrule
\end{tabular}
\end{table}

\section{Discussion}\label{sec:discussion}

\paragraph{Reward saturation hypothesis.}
Our findings are consistent with a \emph{reward saturation} interpretation. SFT teaches the model to compile and formalize specific problems through imitation of ground-truth outputs. When GRPO subsequently trains on the same problems, both components of the dual-stage reward (the compiler gate and the semantic judge) may already be near-saturated, leaving little room for policy improvement.

Training dynamics (\Cref{fig:training_dynamics}) provide additional evidence. The mean reward subplot shows SFT+GRPO-30\% attaining the highest training reward across all variants, yet this model underperforms SFT+GRPO-0\% at evaluation. This disconnect suggests that partial familiarity inflates reward without forcing genuine exploration: the model collects high reward on the ${\sim}$30\% of prompts it has already seen during SFT, while the reward signal on the remaining 70\% is diluted by the inflated baseline. Higher-overlap variants exhibit lower policy entropy throughout training, consistent with the reward-saturation hypothesis that overlap constrains exploration toward memorized solutions rather than novel outputs. Gradient norm and KL loss traces show broadly similar trajectories across variants, suggesting that the overlap effect operates primarily through the reward landscape rather than through optimization instability.

This interpretation aligns with prior evidence that SFT memorizes training distributions while RL generalizes beyond them \citep{chu2025sftmemorizes}: when GRPO data overlaps entirely with SFT, the RL stage may be constrained to the memorized distribution rather than discovering superior policies. From a data-centric perspective, RL's advantage may partly derive from implicit data curation through the reward signal \citep{lu2026datacentric}, and full overlap could eliminate this filtering mechanism. A simpler alternative explanation is that non-overlapping data increases the effective diversity of examples the model encounters across both training stages, and diversity is a well-established driver of generalization in supervised and reinforcement learning alike. Under this view, the overlap effect need not invoke reward saturation at all: fresh GRPO data simply exposes the model to distributional coverage it did not receive during SFT.

\paragraph{Practical implications.}
The compile-semantic gaps observed across all models (\Cref{tab:main_results}, $\Delta$ column) carry a broader methodological implication. Models that achieve the highest compilation rates tend to exhibit the largest gaps, with disparities exceeding 30~pp for the top-compiling baselines. Compilation is necessary but far from sufficient as a quality signal: a model at 84\% C@1 may have fewer than half of those outputs semantically faithful, a disparity entirely invisible under compile-only evaluation. These results motivate dual-metric evaluation as standard practice for autoformalization research. For practitioners, the overlap finding offers an immediately actionable recommendation: partitioning SFT and GRPO data pools is free in terms of compute and consistently improves both metrics.

\section{Limitations}\label{sec:limitations}

\paragraph{Single base model, single run.} All experiments use Qwen3-8B with a single training run per condition. Consequently, we cannot report variance estimates or certify that findings transfer to other model families or scales.

\paragraph{Three overlap levels.} We evaluate 0\%, 30\%, and 100\% overlap. Although the trend is monotonic across these points, three data points are insufficient to characterize a continuous relationship or pinpoint an optimal overlap ratio. Finer-grained sweeps (e.g., 10\%, 50\%, 70\%) would lend greater precision to this finding.

\paragraph{Dataset sizes not independently ablated.} SFT uses 20K pairs and GRPO uses 16K prompts. We do not vary these sizes independently, precluding disentanglement of the overlap effect from potential interactions with dataset scale. A full factorial design (e.g., varying SFT size $\times$ GRPO size $\times$ overlap fraction) would isolate these interactions but requires substantially more training runs.

\paragraph{LLM-based semantic evaluation.} The same Gemini Flash~3 model serves as both the GRPO reward signal during training and the evaluation judge, meaning the evaluation metric is not fully independent of the training objective. We have not validated the judge against formal methods such as BEq \citep{wu2025stepfun} or LeanScorer \citep{zhang2025mathesis}, nor conducted a human-agreement study; such calibration would strengthen confidence in the metric. The 0.7 threshold for S@$k$ is a design choice that affects absolute values but is unlikely to alter relative ordering given the magnitude of observed differences.

\paragraph{Temperature confound.} Internal models use temperature 1.0; external baselines use 0.6 per their model cards. This discrepancy affects absolute comparisons (particularly pass@1) but does not influence the ordering among our overlap variants.

\paragraph{Scope.} This study is limited to Lean~4 with Mathlib. Whether the overlap effect generalizes to other proof assistants (Isabelle, Coq), mathematical domains, or reward formulations remains an open question.

\section{Future Work}\label{sec:future}

\paragraph{Answer injection and chain-of-thought reasoning.} Our answer injection pipeline opens the door to investigating how reasoning traces affect semantic quality in GRPO training. Models with thinking enabled can leverage answer guidance to focus on formalization rather than computation, potentially enabling RL training with chain-of-thought on existing SFT datasets that were not originally designed for RL. Based on our overlap findings, a promising direction is SFT with distilled reasoning traces followed by GRPO on non-overlapping data.

\paragraph{Formal semantic evaluation.} Substituting the LLM judge with BEq verification would yield sound, deterministic semantic metrics and eliminate a key limitation of this study.

\paragraph{Finer overlap sweeps and scaling.} A denser overlap sweep across multiple base models would clarify whether the monotonic trend holds generally and identify the point of diminishing returns.

\paragraph{Joint variation with data scale.} Our study holds dataset sizes fixed. Co-varying overlap and corpus size would reveal whether the overlap effect persists, amplifies, or attenuates at different scales.

\section{Conclusion}\label{sec:conclusion}

Lower SFT-GRPO data overlap consistently yields higher compilation and semantic accuracy, with full overlap rendering the GRPO stage effectively redundant. Our controlled study across six Qwen3-8B conditions shows that 0\% overlap produces the largest gains (10.4~pp beyond SFT alone on Gaokao), with the effect amplified on harder benchmarks. Practitioners designing SFT+GRPO pipelines should construct disjoint data pools when corpus volume permits, as this incurs no additional computational cost, only a different partitioning strategy. More broadly, treating data overlap as an explicit design decision has measurable impact on post-training efficacy at no additional cost.

\section*{Acknowledgements}

We thank the Lean~4 and Mathlib communities, and the teams behind HERALD, NuminaMath, Leanabell-Prover, Lean Workbook, FormaRL, and Qwen3 for the datasets, methods, and models this work builds on.

\bibliographystyle{plainnat}

\begin{thebibliography}{30}

\bibitem[{DeepSeek-AI}(2025)]{guo2025deepseekr1}
{DeepSeek-AI}.
\newblock {DeepSeek-R1}: Incentivizing reasoning capability in {LLMs} via reinforcement learning.
\newblock \emph{arXiv:2501.12948}, 2025.

\bibitem[Xin et~al.(2025)]{xin2025deepseekprover}
Huajian Xin, Daya Guo, Zhihong Shao, et~al.
\newblock {DeepSeek-Prover-V2}: Advancing formal mathematical reasoning via reinforcement learning for subgoal decomposition.
\newblock \emph{arXiv:2504.21801}, 2025.

\bibitem[Wu et~al.(2022)]{wu2022autoformalization}
Yuhuai Wu, Albert~Q. Jiang, Wenda Li, et~al.
\newblock Autoformalization with large language models.
\newblock In \emph{NeurIPS}, 2022.

\bibitem[Huang et~al.(2025)]{huang2025formalrl}
Yige Huang, Xiaohan Jin, Shengyuan Liang, Peilin Li, and Yuanzhi Liu.
\newblock {FormaRL}: Enhancing autoformalization with no labeled data.
\newblock In \emph{COLM}, 2025. arXiv:2508.18914.

\bibitem[Gao et~al.(2025)]{gao2025herald}
Guoxiong Gao et~al.
\newblock {HERALD}: A natural language annotated {L}ean~4 dataset.
\newblock In \emph{ICLR}, 2025. arXiv:2410.10878.

\bibitem[He et~al.(2024)]{he2024leanworkbook}
Yifan He et~al.
\newblock Lean {W}orkbook: A large-scale {L}ean problem set formalized from natural language math problems.
\newblock In \emph{NeurIPS Datasets and Benchmarks}, 2024.

\bibitem[stoney0062(2025)]{stoney2025leanabell}
stoney0062.
\newblock Leanabell-Prover-Formal-Statement.
\newblock HuggingFace dataset, 2025.

\bibitem[{Qwen Team}(2025)]{qwen2025qwen3}
{Qwen Team}.
\newblock Qwen3 technical report.
\newblock \emph{arXiv:2505.09388}, 2025.

\bibitem[Shao et~al.(2024)]{shao2024deepseekmath}
Zhihong Shao, Peiyi Wang, Qihao Zhu, et~al.
\newblock {DeepSeekMath}: Pushing the limits of mathematical reasoning in open language models.
\newblock \emph{arXiv:2402.03300}, 2024.

\bibitem[de~Moura et~al.(2021)]{demoura2021lean4}
Leonardo de~Moura, Sebastian Ullrich, et~al.
\newblock The {L}ean~4 theorem prover and programming language.
\newblock In \emph{CADE}, 2021.

\bibitem[Wang et~al.(2025)]{wang2025kimina}
Haiming Wang, Huajian Xin, Zhengying Liu, et~al.
\newblock {Kimina-Prover}: Automated theorem proving via large language models.
\newblock \emph{arXiv:2504.11354}, 2025.

\bibitem[Zhang et~al.(2024)]{zhang2024gaokao}
Xiaotian Zhang, Chunyang Li, Yi~Zong, et~al.
\newblock Evaluating the performance of large language models via {GAOKAO} benchmark.
\newblock \emph{arXiv:2305.12474}, 2024.

\bibitem[Tsoukalas et~al.(2024)]{tsoukalas2024putnambench}
George Tsoukalas, Jasper Lee, John Jennings, et~al.
\newblock {PutnamBench}: Evaluating neural theorem-provers on the {P}utnam mathematical competition.
\newblock In \emph{NeurIPS Datasets and Benchmarks}, 2024.

\bibitem[Wu et~al.(2025)]{wu2025stepfun}
Yutong Wu, Di Huang, Ruosi Wan, et~al.
\newblock {StepFun-Formalizer}: Unlocking the autoformalization potential of {LLMs} through knowledge-reasoning fusion.
\newblock \emph{arXiv:2508.04440}, 2025.

\bibitem[Chu et~al.(2025)]{chu2025sftmemorizes}
Tianzhe Chu, Yuexiang Zhai, Jihan Yang, and Sergey Levine.
\newblock {SFT} memorizes, {RL} generalizes: A comparative study of foundation model post-training.
\newblock \emph{arXiv:2501.17161}, 2025.

\bibitem[Lu et~al.(2026)]{lu2026datacentric}
Wangyue Lu et~al.
\newblock A data-centric perspective on {RLHF} for code generation.
\newblock \emph{arXiv preprint}, 2026.

\bibitem[Zhang et~al.(2025)]{zhang2025mathesis}
Xiaohan Zhang et~al.
\newblock {Mathesis}: Towards formal theorem proving from natural language.
\newblock In \emph{ICLR}, 2025.

\bibitem[Weng et~al.(2025)]{weng2025autoformalization}
Jianqiao Weng, Yihan Geng, Jinan Xu, Jian Yang, and Yue Zhang.
\newblock Autoformalization of mathematical statements: A comprehensive survey.
\newblock \emph{arXiv:2505.23486}, 2025.

\bibitem[Lin et~al.(2025)]{lin2025goedelprover}
Yong Lin et~al.
\newblock {Goedel-Prover}: A frontier model for open-source automated theorem proving.
\newblock \emph{arXiv:2502.07640}, 2025.

\bibitem[Li et~al.(2024)]{li2024numinamath}
Jia Li, Edward Beeching, Lewis Tunstall, et~al.
\newblock {NuminaMath}: The largest public dataset in {AI4Maths} with 860K pairs of competition math problems and solutions.
\newblock HuggingFace repository, 2024.

\end{thebibliography}

\appendix

\section{Full Compile Pass@\texorpdfstring{$k$}{k} Results}\label{app:full_passk}

\begin{table}[H]
\centering
\small
\caption{Full compile pass@$k$ results on Gaokao-Formal and PutnamBench ($n{=}8$ rollouts).}
\label{tab:full_passk}
\begin{tabular}{@{}l cccc c cccc@{}}
\toprule
& \multicolumn{4}{c}{\textbf{Gaokao-Formal}} && \multicolumn{4}{c}{\textbf{PutnamBench}} \\
\cmidrule(lr){2-5} \cmidrule(lr){7-10}
\textbf{Model} & @1 & @2 & @4 & @8 && @1 & @2 & @4 & @8 \\
\midrule
Base & 19.9 & 26.4 & 33.2 & 40.2 && 11.3 & 16.6 & 22.7 & 29.3 \\
SFT & 61.8 & 75.8 & 86.1 & 92.3 && 28.5 & 40.2 & 52.9 & 65.6 \\
GRPO-only & 50.9 & 58.3 & 64.2 & 69.3 && 36.1 & 44.3 & 52.2 & 60.1 \\
\addlinespace[2pt]
SFT+GRPO-0\% & 77.6 & 86.2 & 91.0 & 93.9 && 47.9 & 59.6 & 69.7 & 78.1 \\
SFT+GRPO-30\% & 76.4 & 84.7 & 89.8 & 92.9 && 46.4 & 56.7 & 65.5 & 71.9 \\
SFT+GRPO-100\% & 62.9 & 76.7 & 86.4 & 92.7 && 29.1 & 40.8 & 53.2 & 65.3 \\
\addlinespace[2pt]
Kimina-7B & 84.2 & 90.9 & 94.7 & 97.2 && 53.5 & 66.4 & 77.1 & 85.7 \\
Mathesis-7B & 84.1 & 89.3 & 92.8 & 95.2 && 63.0 & 74.3 & 82.9 & 88.8 \\
\bottomrule
\end{tabular}
\end{table}

\section{Full Semantic Pass@\texorpdfstring{$k$}{k} Results}\label{app:full_semantic}

\begin{table}[H]
\centering
\small
\caption{Semantic pass@$k$ (S@$k$) across benchmarks ($n{=}8$ rollouts).}
\label{tab:full_semantic}
\begin{tabular}{@{}l cc c cc@{}}
\toprule
& \multicolumn{2}{c}{\textbf{Gaokao-Formal}} && \multicolumn{2}{c}{\textbf{PutnamBench}} \\
\cmidrule(lr){2-3} \cmidrule(lr){5-6}
\textbf{Model} & S@1 & S@8 && S@1 & S@8 \\
\midrule
Base & 10.2 & 19.4 && 3.3 & 7.7 \\
SFT & 41.0 & 70.9 && 14.3 & 34.2 \\
GRPO-only & 28.1 & 40.2 && 11.9 & 19.2 \\
\addlinespace[2pt]
SFT+GRPO-0\% & 51.4 & 72.7 && 23.6 & 43.0 \\
SFT+GRPO-30\% & 48.6 & 70.7 && 22.9 & 38.8 \\
SFT+GRPO-100\% & 40.6 & 69.9 && 14.7 & 34.8 \\
\addlinespace[2pt]
Kimina-7B & 44.6 & 68.1 && 36.8 & 65.3 \\
Mathesis-7B & 49.8 & 71.1 && 43.0 & 69.3 \\
\bottomrule
\end{tabular}
\end{table}

\begin{table}[H]
\centering
\small
\caption{Compile-semantic gap (C@1$-$S@1) across benchmarks.}
\label{tab:cs_gap_full}
\begin{tabular}{@{}l cc@{}}
\toprule
\textbf{Model} & \textbf{Gaokao} & \textbf{Putnam} \\
\midrule
Base & 9.7 & 8.0 \\
SFT & 20.8 & 14.2 \\
GRPO-only & 22.8 & 24.2 \\
SFT+GRPO-0\% & 26.2 & 24.3 \\
SFT+GRPO-30\% & 27.8 & 23.5 \\
SFT+GRPO-100\% & 22.3 & 14.4 \\
\addlinespace[2pt]
Kimina-7B & 39.6 & 16.7 \\
Mathesis-7B & 34.3 & 20.0 \\
\bottomrule
\end{tabular}
\end{table}

\section{Training Configurations}\label{app:configs}

\begin{table}[H]
\centering
\small
\caption{Full SFT configuration (Axolotl).}
\label{tab:sft_config}
\begin{tabular}{@{}ll@{}}
\toprule
\textbf{Parameter} & \textbf{Value} \\
\midrule
Base model & Qwen/Qwen3-8B \\
Chat template & qwen3 (thinking disabled) \\
Sequence length & 4096 \\
Sample packing & Enabled (flex attention) \\
Micro batch / Epochs & 2 / 2 \\
Optimizer & AdamW (fused), cosine LR $2 \times 10^{-5}$, warmup 0.1 \\
Precision & bf16 + tf32 \\
Distributed strategy & FSDP2 (full shard, activation checkpointing) \\
\bottomrule
\end{tabular}
\end{table}

\begin{table}[H]
\centering
\small
\caption{GRPO training configuration (Slime/Megatron-LM).}
\label{tab:grpo_config}
\begin{tabular}{@{}ll@{}}
\toprule
\textbf{Parameter} & \textbf{Value} \\
\midrule
\multicolumn{2}{@{}l}{\textit{Rollout}} \\
\quad Samples per prompt / Batch / Global & 8 / 64 / 128 \\
\quad Max response length & 16{,}384 tokens \\
\quad Temperature / Epochs & 1.0 / 1 \\
\addlinespace[3pt]
\multicolumn{2}{@{}l}{\textit{GRPO}} \\
\quad KL loss coeff / type & 0.01 / low-variance KL \\
\quad Entropy coefficient & 0.001 \\
\quad Clip range ($\epsilon$) / high & 0.2 / 0.28 \\
\quad Reward normalization & Disabled \\
\addlinespace[3pt]
\multicolumn{2}{@{}l}{\textit{Optimizer}} \\
\quad Adam, LR $1 \times 10^{-6}$ (constant) & weight decay 0.1, $\beta_{1,2}$ = 0.9, 0.98 \\
\addlinespace[3pt]
\multicolumn{2}{@{}l}{\textit{Infrastructure}} \\
\quad Actor / Rollout GPUs & 4 (TP=1) / 4 (SGLang, 2/engine) \\
\quad Gradient recomputation & Full (uniform, 1 layer) \\
\quad Checkpoint interval & Every 20 steps \\
\bottomrule
\end{tabular}
\end{table}

\section{System Prompt}\label{app:system_prompt}

\begin{mdframed}[backgroundcolor=codebg, linecolor=codeframe,
  linewidth=0.6pt, roundcorner=2pt,
  frametitle={\small\bfseries System Prompt (shared across SFT, GRPO, and evaluation)}]
{\small\ttfamily\raggedright
You are an assistant that translates natural-language math statements into Lean 4 theorems that compile with Mathlib. The preamble \texttt{import Mathlib} and \texttt{open BigOperators Real Nat Topology Rat} is already provided. Given a math problem or statement, output only a single Lean 4 theorem whose type states the claim, ending the proof with \texttt{:= by sorry}. Do not include any proof steps, explanations, comments, or extra text. Wrap your output in a lean4 code block.\par
}
\end{mdframed}

\section{Semantic Judge Prompt}\label{app:judge_prompt}

\begin{mdframed}[backgroundcolor=codebg, linecolor=codeframe,
  linewidth=0.6pt, roundcorner=2pt,
  frametitle={\small\bfseries LLM-as-Judge Prompt (Gemini Flash 3)}]
{\small\ttfamily\raggedright
You are an expert Lean 4 / Mathlib reviewer. You will be given:\par
\smallskip
1. A ground-truth Lean 4 theorem statement.\par
2. A candidate Lean 4 theorem statement produced by a model.\par
\smallskip
Both compile successfully. Judge how semantically faithful the candidate is to the ground truth, ignoring superficial differences (variable names, open declarations, import style).\par
\smallskip
Score on a scale of 0.0 to 1.0:\par
\smallskip
\hspace{1em}- 1.0 = semantically identical formalization\par
\hspace{1em}- 0.7--0.9 = minor differences that don't change meaning\par
\hspace{1em}- 0.3--0.6 = partially captures the statement but with meaningful errors\par
\hspace{1em}- 0.0--0.2 = wrong or unrelated formalization\par
\smallskip
Respond with ONLY a JSON object: \texttt{\{"score": <float>, "reason": "<brief explanation>"\}}\par
}
\end{mdframed}

\section{Training Dynamics (Raw Traces)}\label{app:training_dynamics}

\begin{figure}[H]
\centering
\includegraphics[width=\textwidth]{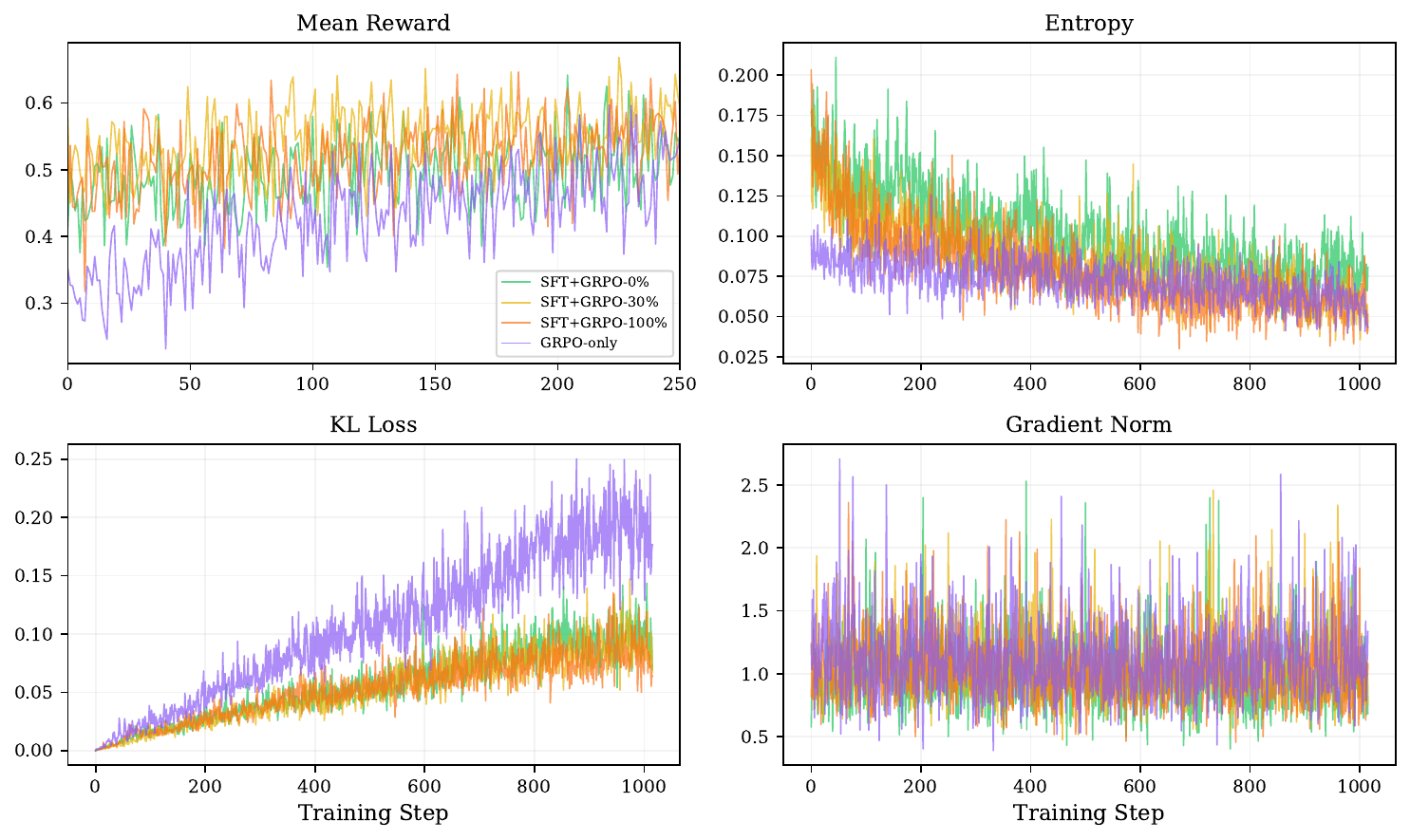}
\caption{Raw (unsmoothed) training dynamics across GRPO variants. Entropy, KL loss, and gradient norm are shown for the full training run.}
\label{fig:training_dynamics_raw}
\end{figure}

\section{Qualitative Error Analysis}\label{app:error_analysis}

We present three examples of model outputs that compile successfully but are semantically incorrect, illustrating distinct failure modes (\Cref{tab:error_summary}).

\begin{table}[H]
\centering
\small
\caption{Summary of qualitative error examples.}
\label{tab:error_summary}
\begin{tabular}{@{}clcc@{}}
\toprule
\textbf{\#} & \textbf{Problem} & \textbf{Failure Mode} & \textbf{Score} \\
\midrule
1 & Putnam 1963 A4 & Wrong relationship & 0.1 \\
2 & Putnam 1964 B4 & Trivially true tautology & 0.0 \\
3 & Putnam 1963 B2 & Dropped constraint & 0.1 \\
\bottomrule
\end{tabular}
\end{table}

\begin{figure}[H]
\begin{mdframed}[backgroundcolor=codebg, linecolor=codeframe, linewidth=0.6pt, roundcorner=2pt]
{\small
\textbf{Example 1: Wrong mathematical relationship (score = 0.1)}\\[4pt]
\textbf{NL problem (Putnam 1963 A4):} Let $\{a_n\}$ be a sequence of positive reals. Show that $\limsup_{n\to\infty}\, n\!\bigl(\frac{1+a_{n+1}}{a_n}-1\bigr) \geq 1$.\\[4pt]
\textbf{Ground truth:} Formalizes the $\limsup \geq 1$ bound and the tightness of the constant~1.\\[4pt]
\textbf{Model output} (compiles):
\begin{lstlisting}
theorem putnam_1963_a4 (a : ℕ → ℝ) (ha : ∀ n, 0 < a n) :
  Filter.Tendsto
    (fun n => n * ((1 + a (n + 1) / a n) - 1))
    Filter.atTop (nhds 1) := by sorry
\end{lstlisting}
\textbf{Judge score: 0.1.} Asserts \emph{convergence} to~1 (\texttt{Tendsto \ldots\ nhds 1}), a strictly stronger (and false) claim compared to the ground truth's $\limsup \geq 1$. Also drops the second part of the problem (that 1 is the tightest such constant).
}
\end{mdframed}
\end{figure}

\begin{figure}[H]
\begin{mdframed}[backgroundcolor=codebg, linecolor=codeframe, linewidth=0.6pt, roundcorner=2pt]
{\small
\textbf{Example 2: Trivially true tautology (score = 0.0)}\\[4pt]
\textbf{NL problem (Putnam 1964 B4):} $n$ great circles on the sphere are in general position. How many regions do they divide the sphere into? Answer: $n^2 - n + 2$.\\[4pt]
\textbf{Ground truth:} Defines great circles, general position, and region counting using sign predicates (20+ lines).\\[4pt]
\textbf{Model output} (compiles):
\begin{lstlisting}
theorem putnam_1964_b4 : ∀ n : ℕ, 0 < n →
  (n ^ 2 - n + 2 : ℕ) = n ^ 2 - n + 2 := by sorry
\end{lstlisting}
\textbf{Judge score: 0.0.} A tautology ($n^2 - n + 2 = n^2 - n + 2$) with zero mathematical content. The model discards the entire geometric setup and produces a trivially provable statement. This illustrates why compilation is a necessary but far from sufficient condition for correctness.
}
\end{mdframed}
\end{figure}

\begin{figure}[H]
\begin{mdframed}[backgroundcolor=codebg, linecolor=codeframe, linewidth=0.6pt, roundcorner=2pt]
{\small
\textbf{Example 3: Weakened statement / dropped constraint (score = 0.1)}\\[4pt]
\textbf{NL problem (Putnam 1963 B2):} Let $S = \{2^m 3^n \mid m, n \in \mathbb{Z}\}$. Show that $S$ is dense in the positive reals.\\[4pt]
\textbf{Ground truth:} Formalizes density as $\text{closure}(S) \supseteq (0, \infty)$.\\[4pt]
\textbf{Model output} (compiles):
\begin{lstlisting}
theorem putnam_1963_b2 (x : ℝ) (hx : x > 0) :
  ∃ m n : ℕ, 2^m * 3^n > x / 2
    ∧ 2^m * 3^n < 2 * x := by sorry
\end{lstlisting}
\textbf{Judge score: 0.1.} Instead of density (arbitrarily close approximation), this claims only that some element of $S$ lies in the interval $(x/2, 2x)$, a vastly weaker statement. Additionally, the model restricts exponents to $\mathbb{N}$ instead of $\mathbb{Z}$, further weakening the claim.
}
\end{mdframed}
\end{figure}

\end{document}